\documentclass{IEEEcsmag}
\usepackage{booktabs} 
\usepackage{amsmath}
\usepackage{amssymb}
\usepackage{bm}
\usepackage{subfigure}
\usepackage{multirow}
\usepackage{graphicx}
\usepackage{verbatim}
\usepackage{algpseudocode}
\usepackage[ruled,linesnumbered,vlined]{algorithm2e}
\newtheorem{myDef}{Definition}[section]
\usepackage[colorlinks,urlcolor=blue,linkcolor=blue,citecolor=blue]{hyperref}

\begin{document}

\sptitle{Department: Head}
\editor{Editor: Name, xxxx@email}

\title{EAGLE: Contrastive Learning for Efficient Graph Anomaly Detection}

\author{Jing Ren}
\affil{Institute of Innovation, Science and Sustainability, Federation University Australia, Ballarat, VIC 3353, Australia}

\author{Mingliang Hou, Zhixuan Liu}
\affil{School of Software, Dalian University of Technology, Dalian 116620, China}

\author{Xiaomei Bai*}
\affil{Computing Center, Anshan Normal University, Anshan 114007, China}

\markboth{Department Head}{Paper title}

\begin{abstract}
Graph anomaly detection is a popular and vital task in various real-world scenarios, which has been studied for several decades. Recently, many studies extending deep learning-based methods have shown preferable performance on graph anomaly detection. However, existing methods are lack of efficiency that is definitely necessary for embedded devices. Towards this end, we propose an Efficient Anomaly detection model on heterogeneous Graphs via contrastive LEarning (EAGLE) by contrasting abnormal nodes with normal ones in terms of their distances to the local context. The proposed method first samples instance pairs on meta path-level for contrastive learning. Then, a graph autoencoder-based model is applied to learn informative node embeddings in an unsupervised way, which will be further combined with the discriminator to predict the anomaly scores of nodes. Experimental results show that EAGLE outperforms the state-of-the-art methods on three heterogeneous network datasets.
\end{abstract}

\maketitle

\chapterinitial{Heterogeneous graphs}, which consist of multiple types of node objects and relationships between the node pairs, have become a popular data structure for representing a wide variety of real-world network datasets~\cite{yu2022graph}. Typical examples include social networks, bibliographic networks, and transportation networks. Recent years have witnessed increasing attention on graph data mining and analysis tasks, such as node/graph classification, recommendation systems, and anomaly detection~\cite{akoglu2015graph}. The goal of anomaly detection tasks is to identify patterns that deviate from other samples in a specific context. It has significant implications in preventing  real-world systems from resulting in huge damage~\cite{ma2021comprehensive}. For instance, a recent study\footnote{\url{https://www.zdnet.com/article/online-fake-news-costing-us-78-billion-globally-each-year/}} revealed that the direct economic cost inflicted by online fake news reached around $\$$78 billion a year globally.

In the last five years, there is significant progress on graph anomaly detection models with the introduction of machine learning techniques~\cite{markovitz2020graph}, which largely reduces the dependency on human experts' domain knowledge. By applying graph neural networks (GNNs) into graph representation learning models, these models are expressive enough to fully support graph anomaly detection, which consequently has shown superior performance on different kinds of graphs and anomaly detection tasks~\cite{fan2020metagraph,liu2022deep}. Specifically, Liu et al. \cite{liu2018heterogeneous} employ a graph convolution network (GCN) on the original account-device graph, and embeds both structure and node attribute information of each vertex into a latent vector space. The proposed algorithm also adaptively estimates the attention coefficients when dealing with different types of subgraphs. As for real-world applications, Wang et al.~\cite{wang2020defending} propose a structured anomaly detection framework to defend Water Treatment Networks (WTNs) by modeling the spatiotemporal characteristics of cyber attacks in WTNs. To solve the problem of graph inconsistency and imbalance that generally existed in the anomaly detection tasks, Zhang et al. \cite{zhang2021fraudre} investigate three aspects of graph inconsistencies and presented a fraud detection model based on Graph Neural Networks. 


However, existing heterogeneous graph anomaly detection still faces several issues: First, due to the shortage of enough ground-truth labels of anomalies in some real-world environments, supervised learning models cannot be applied directly because they are highly dependent on balanced datasets with enough labels for training. Therefore, graph anomaly detection models have to be trained in an unsupervised manner and most graph convolutional models are usually not robust enough~\cite{zhu2019robust}. Second, heterogeneous graphs include complicated structures (multiple nodes and edge types) as well as abundant node attribute information, which increases the difficulty of identifying diverse and complex anomalies. Finally, a major limitation of current deep graph learning approaches is that they lack efficiency. An efficient model is conducive to be set into embedded platforms in anomaly detection tasks.

In light of these limitations and challenges, we propose an Efficient Anomaly detection model on heterogeneous Graphs via contrastive LEarning (EAGLE for abbreviation). Based on the assumption that anomalous nodes tend to be more distant from the local context of the nodes than normal ones, we aim to contrast each node with the local meta paths of the target node. By sampling the meta path-level instance pairs from the heterogeneous network and using them to train the contrastive learning model, the semantics of the heterogeneous network is preserved. In particular, given the node attribute features as input, the learned node embeddings can capture the structure and attribute information simultaneously. The learning objective of EAGLE is to distinguish the elements between the positive and negative instance pairs, and the predicted scores can be used to measure the anomalous degree of nodes. To summarize, the main contributions are as follows:
\begin{itemize}
\item We propose a contrastive self-supervised learning framework, 	EAGLE, for the anomaly detection problem on heterogeneous networks.
	\item We present a novel view to generate positive and negative instances on heterogeneous graphs at the meta path-level, which could comprehensively preserve the rich semantics as a form of transferable knowledge for downstream anomaly detection tasks.
	\item We design an efficient anomaly detection model which can save computation resources and be applied to embedded devices.
	\item We conduct extensive experiments and analysis on three real-world heterogeneous graphs to demonstrate that the proposed model EAGLE significantly outperforms a range of baseline methods.
\end{itemize}
%
%

\section{Related Work}~\label{sec2}
In this section, we introduce two most related research topics: graph anomaly detection, and contrastive learning.

\subsection{Graph Anomaly Detection}
Anomaly detection aims to identify the unusual patterns that significantly deviate from the majority in a dataset, which is a popular and vital task in various research contexts. Considering the capabilities of reconstruction-based models in distinguishing anomalies from normal graph data, a series of autoencoder-based graph anomaly detection methods have been proposed, which is also the main module used in this paper. For example, DOMINANT \cite{ding2019deep} aims to detect node anomalies by a deep autoencoder that reconstructs the original data. Similarly, AnomalyDAE \cite{fan2020anomalydae} is also a deep representation learning framework for anomaly detection on attributed networks through a dual autoencoder.

However, existing heterogeneous graph anomaly detection approaches are mostly focused on a specific context or application, which lack generalization. By integrating contrastive learning into a deep graph autoencoder, we will develop a novel algorithm for heterogeneous network anomaly detection without any limits on application scenarios or expert knowledge.

\subsection{Contrastive Learning}
Self-supervised learning models aim to learn graph representations from unlabeled data by solving a series of pretext tasks. As a bunch of self-supervised learning, contrastive learning is built on the idea of learning common features between positive instances and distinguish the differences between positive and negative instances. Hence, the target task of contrastive learning models is to find an appropriate way for generating positive and negative data samples.

To make full use of the unlabeled graph-structured data, some recent efforts have been put into graph contrastive learning models~\cite{qiu2020gcc}. Among most existing graph contrastive learning methods, the common data augmentation schemes includes uniformly dropping edges/nodes, or randomly shuffling node features. CPT-HG~\cite{jiang2021contrastive} designs contrastive pre-training strategies for heterogeneous graphs at relation- and metagraph-levels. Zhu et al.~\cite{zhu2021graph} design two data augmentation strategies from two perspectives. However, most of these methods focus on homogeneous network, and are applied into common downstream tasks. Considering the specialty of anomaly detection tasks, this paper proposes a novel way for heterogeneous graph data augmentations to learn abundant structure information more effectively.

\begin{figure}[htbp]
	\centering
	\includegraphics[width=0.5\textwidth]{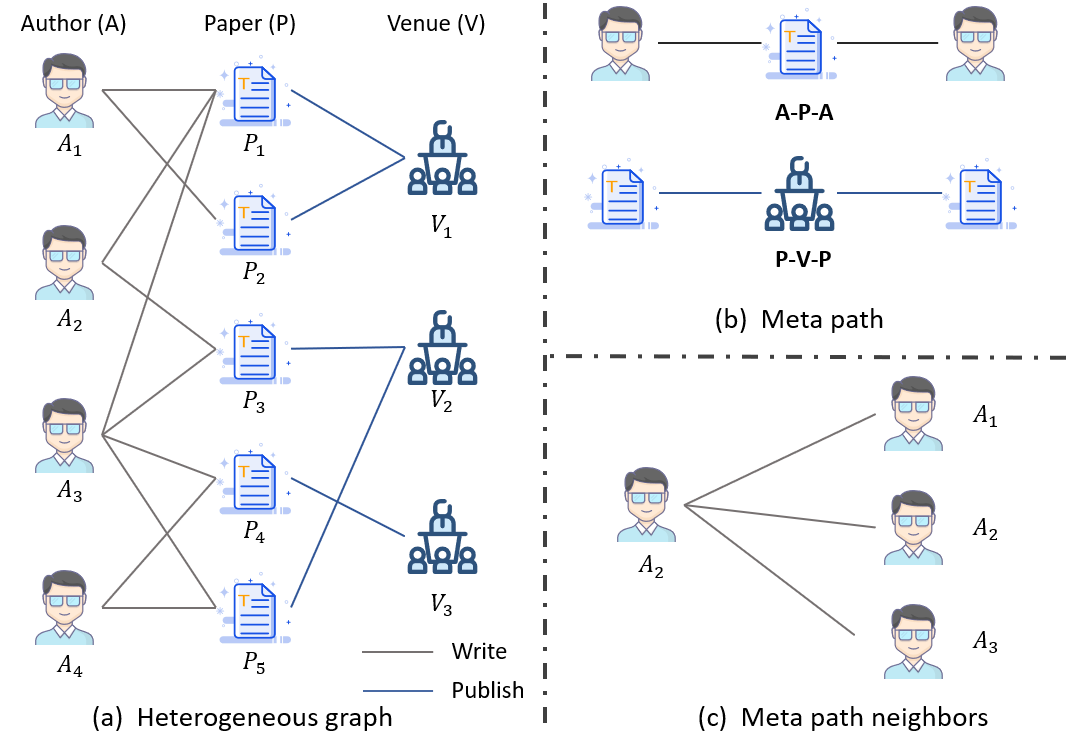}\\
	\caption{An example of heterogeneous graph, meta-path, and meta-path neighbors. (a) There are three types of nodes, namely Author, Paper, and Venue, and two types of edge, namely Write and Publish in this heterogeneous graph. (b) shows two kinds of meta paths, i.e., Author-Paper-Author, and Paper-Venue-Paper. (c) lists all neighbors of node $A_2$ according to the meta path A-P-A, noting that node $A_2$ itself is also regarded as one of its neighbors. }
	\label{fig:hegraph}
\end{figure}
\section{Preliminaries}\label{sec3}
In this section, we give some formal definitions relevant to heterogeneous graphs and formalize the problem of heterogeneous graph anomaly detection.

\begin{myDef}[Heterogeneous Graph]\label{def:Hegraph}
	A heterogeneous graph (a.k.a., heterogeneous information
	networks) is defined as $\mathcal{G}=(\mathcal{V,E},$ 	$\mathcal{A,R},\phi,\varphi)$, where $\mathcal{V}$ and $\mathcal{E}$ denote the node set and edge set, respectively. $\mathcal{A}$ and $\mathcal{R}$ refer to the sets of node types and edge types. Here, $\phi:\mathcal{V} \to \mathcal{A}$ is the node type mapping function, and $\varphi:\mathcal{E} \to \mathcal{R}$ is the edge type mapping function, where $|\mathcal{A}|+|\mathcal{R}|>2$.
\end{myDef}

A toy example of a heterogeneous graph is shown in Figure~\ref{fig:hegraph} (a).

\begin{myDef}[Meta path]
	\label{def:metapath}
	A meta-path $p\in \mathcal{P}$ is defined on the network schema in the form of $A_1\stackrel{R1}{\longrightarrow}A_2\stackrel{R2}{\longrightarrow}...\stackrel{Rl}{\longrightarrow}A_{l+1}$, which could be abbreviated as $A_1A_2...A_{l+1}$. A metapath is composed of a composite relation $R=R_1\circ R_2\circ \dots R_l$ ranging from node types $A_1$ to $A_{l+1}$, where $\circ$ refers to the composition operator on relations.
\end{myDef}

As a basic analysis tool for heterogeneous graphs, a meta-path captures the proximity among multiple nodes from a specific semantic perspective, which could be seen as a high-order structure (shown in Figure~\ref{fig:hegraph} (b)). For example, the meta path ``Author-Paper-Author" $(APA)$ describes that two authors collaborated on a particular paper, and ``Paper-Venue-Paper" $(PVP)$ indicates that two papers are published in the same venue (conference or journal).

\begin{myDef}[Meta path neighbors]
	\label{def:metapathnei}
	For a node $v_i$ and a meta path $p\in \mathcal{P}$, meta path neighbors $\mathcal{N}_{v_i}^p$ is defined as the nodes which linked with node $v_i$ through meta-path $p$.
\end{myDef}
The meta-path-based neighbors can exploit more structure information in a heterogeneous graph.

The problem of anomaly detection on heterogeneous graphs is defined as:
\begin{myDef}[AD on Heterogeneous Graphs]
	\label{def:HegraphAD}
	Provided a heterogeneous graph $\mathcal{G}=(\mathcal{V,E},\mathbf{X})$, the goal is to learn an anomaly score function $\mathit{F}( \cdot)$ to calculate the anomaly score of each node. The anomaly score measuring the abnormality degree of the node is ranked in descending order. Then, the node whose anomalous score is greater than a threshold in this ranking list will be regarded as anomalous.
\end{myDef}

In the paper, we consider the setting of unsupervised anomaly detection without relying on any labeling information, and the strategies follow a two-step paradigm, namely pre-training and fine-tuning. Specifically, we pre-train the model on a part of a large-scale heterogeneous graph while fine-tuning the anomaly detection task on the remaining part.

\section{The Proposed Model: EAGLE}~\label{sec4}
\begin{figure*}[htbp]
	\centering
	\includegraphics[width=0.9\textwidth]{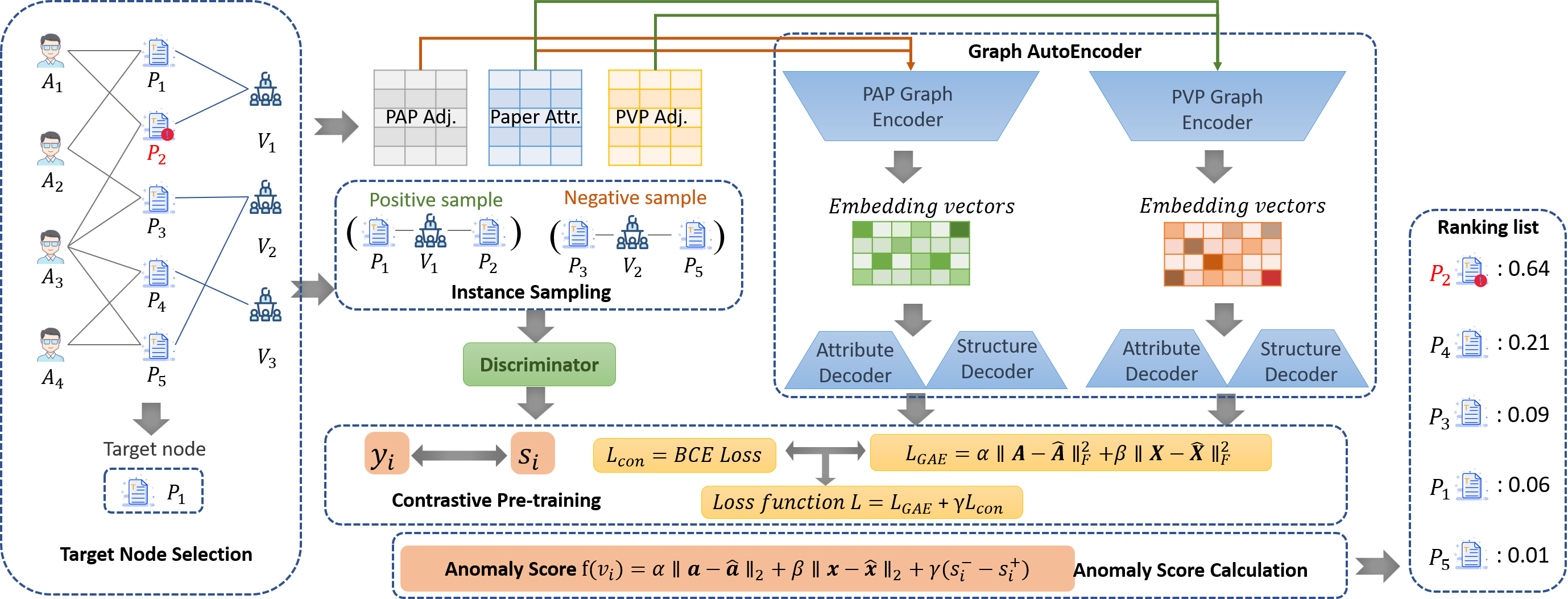}\\
	\caption{The conceptual framework of EAGLE is composed of four main components: target node selection, instance sampling, contrastive pre-training for graph representation learning, and anomaly score calculation. In this figure, the adjacency matrix PAP and PVP Adj. derived by Paper-Author-Paper and Paper-Venue-Paper meta-paths and the paper attribute matrix Paper Attr. are input to the graph autoencoder for graph representation learning. The process of instance pair sampling is to select positive meta path and negative meta path based on the situation that whether the target node is located in the meta path. The Discriminator aims to distinguish the negative pair from the positive pairs. Finally, the discrimination score and the reconstruction error of the graph decoder are combined to calculate the anomaly score.}
	\label{fig:eagle}
\end{figure*}

In this section, we present a framework, namely EAGLE, based on contrastive pre-training strategy for heterogeneous graph anomaly detection. As shown in Figure~\ref{fig:eagle}, EAGLE mainly consists of four different components, i.e.,  target node selection, instance sampling, contrastive pre-training for graph representation learning, and anomaly score calculation. After selecting the target node in a heterogeneous graph, we perform data augmentations on the meta path-level. The latent representations for these positive and negative instances are encoded by a GAE-based contrastive pre-training model. Then, the discriminative score for each instance pair calculated by a discriminator is deployed to be further combined with the reconstruction error to calculate the node anomaly score. Finally, the anomalies could be detected by ranking the anomaly scores.

\subsection{Instance Pair Sampling}
In most graph contrastive learning models, the operations of data augmentation include nodes/edges adding/dropping, attribute masking, and subgraph sampling. Existing research has verified that the design of data augmentation plays a critical role in graph representation learning. According to the definition of anomaly detection task that an anomaly is usually referred to the inconsistency between it and its surroundings, we design a new way to generate positive and negative samples, which could capture such local structures. To learn the comprehensive semantics ingrained in the heterogeneous graph, a common way is to employ meta paths to explore high-order relations. Therefore, we generate meta-path instances to construct the positive samples and negative examples. Specifically, the instance pair on meta path-level is ``target node v.s. meta path".

\textbf{Positive Samples} Given a meta path $p\in\mathcal{P}$ and target node $u$, a positive meta path is defined as a set of nodes $\mathcal{I}(p)$ containing node $u$ on the network schema. For example, in Figure~\ref{fig:eagle}, the positive meta path $P_1-V_1-P_2$ ($P_1V_1P_2$ for brevity) for node $P_1$ originates from $P_1$ and a set of nodes $\mathcal{I}(p)=\{P_1,V_1,P_2\}$. Then, the positive samples from meta paths $w.r.t.$ node $u$ is defined as:
\begin{equation}
\mathcal{P}_u^{mepa}=\bigcup\limits_{p\in\mathcal{P}}\mathcal{I}(p),
\end{equation}
where $\mathcal{P}$ is the set of pre-defined meta paths.

\textbf{Negative Samples} In contrast to positive samples from meta paths $w.r.t.$ the target node $u$, negative samples from the meta path level are defined as the meta paths without node $u$. Considering that if a meta path is far away from node u, it is easily distinguished from the positive pairs by the discriminator. Therefore, we limit the range of the negative meta-paths. Specifically, at least one node in the negative meta paths should be the direct neighbor of the target node $u$. Given a meta path $p\in\mathcal{P}$ and target node $u$, a negative meta path is defined as a set of nodes $\mathcal{I}(p)$ without node $u$ on the network schema. For example, in Figure~\ref{fig:eagle}, the negative meta path $P_3-V_2-P_5$ for node $P_1$ contains a set of nodes $\mathcal{I}(p)=\{P_3,V_2,P_5\}$. Then, the negative samples from meta paths $w.r.t.$ node $u$ is defined as:
\begin{equation}
\mathcal{N}_u=\bigcup\limits_{p\in\mathcal{P}}\mathcal{I}(p)\setminus\{u\},
\end{equation}
where $\setminus $ means node $u$ is excluded from $\mathcal{I}(p)$.
\subsection{Graph AutoEncoder-based Contrastive Learning Model}
As shown in Figure~\ref{fig:eagle}, our proposed GNN-based contrastive learning model mainly consists of two components: Graph AutoEncoder (GAE) for network representation learning and a discriminator for instance pair contrast. Despite that there are two encoders and four decoders in our framework, the internal framework of the corresponding modules is the same by inputting different adjacency matrices according to the metapath neighbors.
\subsubsection{Attributed Graph Encoder}
Deep AutoEncoder is a traditional unsupervised model for representation learning among different kinds of deep neural networks. As one of the reconstruction-based models, deep AutoEncoder has achieved remarkable performance on anomaly detection tasks as well. Despite deep GAE has been applied into attributed networks, applying GAE to learn more informative node representations on heterogeneous graph remains a daunting task. To this end, we propose a new structure of GAE by combining two encoders with graph convolutional networks (GCN) simultaneously, which can incorporate both paper-author and paper-venue relationships into node representation learning.

Mathematically, GCN learns node representations in a layer-wise way:
\begin{equation}
\mathbf{H}^{(l+1)}=f(\mathbf{H}^{(l)},\mathbf{A}|\mathbf{W}^{(l)}),
\end{equation}
where $\mathbf{H}^{(l)}$ is the node embeddings in layer $l$, and $\mathbf{W}^{(l)}$ is the trainable weight parameter matrix in layer $l$. The initial input to the first layer of the model $\mathbf{H}^{(0)}$ is the attribute matrix $\mathbf{X}$, and $\mathbf{A}$ is the adjacency matrix. The function of the GCN in each layer can be denoted as:
\begin{equation}
f(\mathbf{H}^{(l)},\mathbf{A}|\mathbf{W}^{(l)})=\sigma(\widetilde{\mathbf{D}}^{-\frac{1}{2}}\widetilde{\mathbf{A}}\widetilde{\mathbf{D}}^{-\frac{1}{2}}\mathbf{H}^{(l)}\mathbf{W}^{(l)}),
\end{equation}
where $\widetilde{\mathbf{A}}=\mathbf{I}+\mathbf{A}$ and $\widetilde{\mathbf{D}}$ is a digonal matrix of $\widetilde{\mathbf{A}}$. By setting the attribute matrix as the model input, the structure information of nodes are learned by stacking convolutional layers iteratively. Therefore, our encoder could learn both node structure and attribute information simultaneously.
\subsubsection{Attribute Reconstruction Decoder}
is designed to reconstruct the node attribute information by minimizing the reconstruction errors. The structure of attribute decoder is also a graph convolutional layer to reconstruct the input node attributes according to the learned representation $\mathbf{H}$:
\begin{equation}
\widehat{\mathbf{X}}=f_{Relu}(\mathbf{H},\mathbf{A}|\mathbf{W}).
\end{equation}
$\widehat{\mathbf{X}}$ is the reconstructed attribute matrix and the reconstructed errors $\widehat{\mathbf{X}}-\mathbf{X}$ could be utilized as a part of calculating anomaly scores.
\subsubsection{Structure Reconstruction Decoder}
Similar to the attribute reconstruction decoder, structure reconstruction decoder is applied to reconstruct the structure information of the input graph according to the latent embeddings  $\mathbf{H = \{h_1,h_2,\dots,h_N\} }\in \mathbb{R}^{N\times d}$ learned by the encoder. The structure information is formulated as the adjacency matrix of the graph, so the aim of the structure reconstruction decoder is to predict whether node $i$ is a meta-path neighbor of node $j$:
\begin{equation}
p(\widehat{\mathbf{A}}_{i,j}=1|\mathbf{h}_i,\mathbf{h}_j)=sigmod(\mathbf{h}_i,\mathbf{h}_{j}^{T}).
\end{equation}
$\widehat{\mathbf{A}}$ is the reconstructed adjacency matrix, and the structure reconstruction error $\widehat{\mathbf{A}}-\mathbf{A}$ can also be used as a part of node anomaly score calculation.
\subsubsection{Discriminator}
After obtaining the node and instance embeddings from the GAE-based module, a discriminator module is applied to contrast the embeddings of the two elements in an instance pair and finally predicts the discrimination scores for positive pairs and negative pairs. The design of the discriminator is based on the bilinear transformation~\cite{liu2021anomaly}, where the discrimination score is calculated by:

\begin{equation}
s_i=\mathcal{D}(\mathbf{h}_i,\mathbf{h}_i^{tn})=\sigma\left(\mathbf{h}_i^{tn\mathrm{T}}\mathbf{W}\mathbf{h}_i\right),
\end{equation}
where $\mathbf{W}$ is the weight matrix of the discriminator, $\mathbf{h}_i$ and $\mathbf{h}_i^{tn}$ are the selected instance and target node embeddings.
\subsubsection{Objective Function}
Due to the characteristics of anomaly detection tasks, our objective is to make the predicted discrimination score $s_i$ and the ground-truth label $y_i$ as close as possible. Here, we form our contrastive objectives based on standard binary cross-entropy (BCE) loss, which has been validated by other contrastive learning models:

\begin{equation}
L_{con}=-\sum_{i=1}^{N}y_i\mathrm{log}(s_i)+(1-y_i)\mathrm{log}(1-s_i),
\end{equation}
where $L_{con}$ denotes the contrastive loss across all nodes in the graph. The objective function of a deep graph autoencoder can be formulated as:
\begin{equation}
L_{GAE}=\alpha \parallel \mathbf{A}-\widehat{\mathbf{A}} \parallel_{F}^{2} +\beta \parallel \mathbf{X}-\widehat{\mathbf{X}} \parallel_{F}^{2},
\end{equation}
where $\alpha$, and $\beta$ are controlling parameters that balance the impacts of structure reconstruction error and attribute reconstruction error.
To jointly learn the contrastive learning and reconstruction errors, the objective function of our proposed deep graph contrastive autoencoder can be formulated as:
\begin{equation}
L=L_{GAE}+\gamma L_{con},
\end{equation}
where $\gamma$ is a controlling parameter that balances the impacts of the contrastive learning model.

\subsection{Anomaly Score Computation}
With the GAE-based contrastive learning model mentioned above, the proposed model could learn robust and informative embeddings including both attribute and structure information. Experimental results have shown that normal input nodes could be reconstructed with less error than abnormal nodes~\cite{ding2019deep}. Therefore, the reconstruction error could be used as a part of anomaly score calculation process. As for the discrimination score, the discriminator will output close to 0 for negative pair $s^{-}$ and 1 for positive pair $s^{+}$ of a normal node, while the output score is close to 0.5 for both positive and negative sample of an anomalous node. So the anomaly score $s_i^{-}-s_i^{+}$ of an anomalous node is close to 0 while a normal node is close to -1.
Then, the anomaly score of node $v_i$ is calculated as:
\begin{equation}
f(v_i)=\alpha \parallel \mathbf{a}-\widehat{\mathbf{a}} \parallel_{2} +\beta \parallel \mathbf{x}-\widehat{\mathbf{x}} \parallel_{2}+ \gamma (s_i^{-}-s_i^{+}).
\end{equation}
Therefore, the predicted scores for abnormal nodes are larger than normal nodes.

\subsection{Model Analysis}
Here we give the analysis of the proposed EAGLE as follows:
\begin{itemize}
	\item EAGLE could learn informative embeddings for nodes in a heterogeneous network, which incorporates both attribute and structure features into the learning process.
	\item EAGLE is highly efficient and can be easily generalized into large-scale networks through pre-training procedure. For instance pair sampling, we suppose the number of instance samples on the meta path-level is $P$. The time complexity of instance pair sampling for a node is $\mathcal{O}(P)$. The time complexity of learning node representations with a GAE-based model is $\mathcal{O}(edf)$, where $e$ is the number of non-zero elements in the adjacency matrix. $d$ and $f$ are features and weight matrix dimensions. Then, the time complexity of EAGLE is $\mathcal{O}(edFP)$, where F is the	sum of the weight dimension for all layers.
	\item EAGLE is trained with self-supervised information in an end-to-end manner, without requiring a large volume and variety of labeled data for downstream tasks.
\end{itemize}
%
\section{Experiments}~\label{sec5}
In this section, we conduct experiments on three real-world heterogeneous network datasets to demonstrate the effectiveness of the proposed EAGLE model. Dataset and experimental setup are first introduced for better understanding. Then, the experiments cover the anomaly detection performance and time efficiency comparison with SOTA, and parameter study.
\subsection{Dataset}
Three heterogeneous graph datasets, DBLP\footnote{\url{https://dblp.uni-trier.de/}}, Aminer\footnote{\url{https://www.aminer.cn/oag-2-1}}~\cite{tang2008arnetminer}, and Yelp\footnote{\url{https://www.yelp.com/dataset}}, are selected to evaluate our method. For DBLP and Aminer, each author, paper, and venue are initialized by a bag-of-words representation of their paper abstracts. For Yelp, the input feature vector user, business, and review are initialized according to their corresponding reviews.

Considering that there is no ground truth label for anomalies in these datasets, we inject synthetic anomalies into the datasets for evaluation. Specifically, we inject contextual anomalies based on the assumption that the node attribute embedding deviating from its neighbors is regarded as anomalous. Concretely, we select a target node and randomly select $k$ nodes. After calculating the Euclidean distance between the target node and these $k$ nodes, we find the node (from these $k$ nodes) that has the largest Euclidean value to the target node. Then, we replace the attribute embedding of the target node with the attribute embedding of this node. Here, we fix $k=50$ for all kinds of datasets. The details of the three datasets are summarized in Table~\ref{tab:dataset}.

\begin{table}[htbp]
	\centering
	\caption{STATISTICS OF THE DATASETS. DBLP AND AMINER ARE CITATION NETWORKS. YELP IS A SOCIAL NETWORK.}
	\resizebox{0.47\textwidth}{!}{
		\begin{tabular}{c||cc|cc|c}
			\toprule
			Dataset & Node type   & Nodes & Edge type & Edges & Anomalies\\
			\midrule
			\multirow{3}{*}{DBLP} & Author (A)   & 10223 & 	\multirow{2}{*}{A-P} &  \multirow{2}{*}{13119}&  \multirow{3}{*}{150}\\
			& Paper (P)  & 3596 &\multirow{2}{*}{P-V}  &\multirow{2}{*}{3596}  \\
			& Venue  (V) & 456 &  &  \\
			\midrule
			\multirow{3}{*}{Aminer} & Author (A) & 19938 &  \multirow{2}{*}{A-P}   & \multirow{2}{*}{21944}&  \multirow{3}{*}{300} \\
			& Paper (P) & 7612& \multirow{2}{*}{P-V}  & \multirow{2}{*}{7612}  \\
			& Venue (V) & 857 &  &  \\
			
			\midrule
			\multirow{3}{*}{Yelp} & User (U) & 2894 & \multirow{2}{*}{U-B} &\multirow{2}{*}{12456}&  \multirow{3}{*}{450}\\
			& Business (B)& 10272 & \multirow{2}{*}{B-R} & \multirow{2}{*}{12878} \\
			& Review (R) & 12878 &  & \\
			\bottomrule
		\end{tabular}%
		\label{tab:dataset}}%
\end{table}%

\subsection{Experimental Setup}
\subsubsection{Baselines}
\begin{itemize}
	
	\item \textbf{AnomalyDAE} \cite{fan2020anomalydae} is a deep joint representation learning framework for anomaly detection on attributed networks through a dual autoencoder.

	\item \textbf{HeGAN} \cite{hu2019adversarial} employs adversarial learning for Heterogeneous Information Network embedding.
	\item \textbf{DGI} \cite{velickovic2019deep} is an approach for learning unsupervised representations on graph-structured data by maximizing local mutual information among the graph embeddings.
	\item \textbf{DOMINANT} \cite{ding2019deep} aims to detect node anomalies by a deep autoencoder that reconstructs the original data.
	\item \textbf{FRAUDRE} \cite{zhang2021fraudre} presents a fraud detection model based on Graph Neural Networks.
\end{itemize}
\subsubsection{Pre-training and Fine-tuning Setting}
The graph autoencoder model is pre-trained and the initialized parameters of the model could be directly utilized into downstream tasks, i.e., anomaly detection in this work. The fine-tuning process will then be applied according to the specific anomaly detection tasks and be utilized to compare the model performance. Intuitively, fine-tuning on a learned framework will cost less running time than the randomly initialized parameters. For each dataset in this work, we randomly split the whole graph into two graphs for pre-training and fine-tuning. Specifically, the percentage of pre-training dataset is 30\% in DBLP and Aminer, and 70\% percentage of pre-training dataset are used in Yelp.

\subsubsection{Metrics}
To quantitatively evaluate the performance of our model on heterogeneous network anomaly detection, We use AUC, a widely applied metric for anomaly detection in previous works. AUC measures the area under the ROC curve ranging from 0 to 1. A larger value of AUC indicates better detection performance. The ROC curve is defined as the true positive rate (an anomaly is recognized as an anomaly) against the false positive rate (a normal node is recognized as an anomaly).
\subsubsection{Parameter Settings}
We implement our EAGLE with Tensorflow 2.0 and adopt the Adam optimizer to train the proposed model. For the proposed EAGLE on all three datasets, we set the learning rate to 0.001, 0.006, 0.001 for the three datasets, respectively, the embedding dimension is fixed to be 64, and the hyper-parameters $\alpha$, $\beta$, and $\gamma$ are set to 0.8, 0.2 and 0.3.

\begin{center}
	\begin{table*}[htbp]
		\caption{AUC VALUES AND TIME EFFICIENCY COMPARISON ON THREE BENCHMARK DATASETS. THE BEST PERFORMING METHOD IN EACH EXPERIMENT IS IN BOLD.}
		\label{table}
		\centering
		\renewcommand\arraystretch{1.3} 
		\begin{tabular}{c|ccc|ccc}
			\toprule[1.5pt] 
			& \multicolumn{3}{|c|}{AUC}                                         & \multicolumn{3}{c}{Time (s)}                         \\ \midrule[1pt] 
			\makebox[0.2\textwidth][c]{Methods}    & \makebox[0.1\textwidth][c]{DBLP}            & \makebox[0.1\textwidth][c]{Aminer}          & \makebox[0.1\textwidth][c]{Yelp}            & \makebox[0.1\textwidth][c]{DBLP}            & \makebox[0.1\textwidth][c]{Aminer}          & \makebox[0.1\textwidth][c]{Yelp}            \\ \midrule[1pt]  
			AnomalyDAE & 0.8018          & 0.8601          & 0.8671          & 1.0016          & 7.2981          & 11.3963         \\
			HeGAN      & 0.5534          & 0.6590          & 0.8502          & 10 min+         & 30 min+         & 30 min+         \\
			DGI        & 0.5372          & 0.5669          & 0.6008          & 2.1687          & 8.9925          & 9.5246          \\
			DOMINANT   & 0.7605          & 0.8302          & 0.8011          & 0.3096          & 0.9455          & 5.6728          \\
			FRAUDRE    & 0.6986          & 0.8167          & 0.8668          & 1.0529          & 1.8536          & 3.6299          \\
			EAGLE\_Pre & 0.9285          & 0.9551          & 0.9749          & 0.1651          & 0.7876          & 2.0280          \\
			EAGLE      & \textbf{0.9502} & \textbf{0.9592} & \textbf{0.9826} & \textbf{0.0984} & \textbf{0.4124} & \textbf{0.2149} \\ \bottomrule[1.2pt] 
		\end{tabular}
	\label{tab:comparison}
	\end{table*}
\end{center}
\begin{table}[htbp]
	\centering
	\caption{EFFECT OF DIFFERENT READOUT FUNCTION ON AUC VALUES. THE BEST RESULTS IN EACH FUNCTION ARE IN BOLD.}
	\resizebox{0.47\textwidth}{!}{
		\begin{tabular}{c|ccc}
			\toprule
			& Aminer   & DBLP & Yelp\\
			\midrule
			Max Pooling & \textbf{0.9619}   & 0.9409 & 	0.9766 \\
			Min Pooling & 0.9617   & 0.9425 & 	0.9756 \\
			\midrule
			EAGLE (Average Pooling)	 & 0.9592   & \textbf{0.9502} & 	\textbf{0.9826} \\
			\bottomrule
		\end{tabular}%
		\label{tab:readout}}%
\end{table}%
\subsection{Comparison with the State-of-the-art Methods}
In this subsection, we evaluate the anomaly detection performance and running time of EAGLE by comparing it with the baseline models. The comparison of AUC values and running time are shown in Table~\ref{tab:comparison}. According to the results, we make the following observations:
\begin{itemize}
	\item The proposed deep model EAGLE outperforms other baseline methods on all three heterogeneous networks. Specifically, EAGLE has a significant improvement of 28.4\% on average compared with other deep learning methods. These results could demonstrate that EAGLE could learn informative embeddings for nodes, which incorporates both attribute and structure features into the learning process.
	\item Compared with the end-to-end anomaly detection methods, the graph representation learning methods, DGI and HeGAN, show poorer anomaly detection performance. The main reason is that graph representation methods do not have appropriate calculation formulas for detecting node anomalies.
	\item Despite the fact that DOMINANT is a deep graph autoencoder-based method, EAGLE can have about 15\% improvements on three benchmark datasets. This is because EAGLE combines contrastive learning mechanisms into node representation learning and incorporates the pre-training stage before training.
	\item EAGLE\_Pre shown in the table refers to the proposed model EAGLE without a pre-training process, which has a lower AUC value and higher running time. This result proves the efficiency and effectiveness of the pre-training process.
\end{itemize}
\subsection{Parameter Study}

\subsubsection{Effect of Readout Function}
Considering that there are several types of readout functions in our framework, we compared the impact of three common pooling functions on AUC values, namely Min Pooling, Max Pooling, and Average Pooling. The process of Min/Max pooling is to select the minimum/maximum value from all embedding dimensions, while Average Pooling refers to the process of calculating the average values of all embeddings. EAGLE chooses the Average Pooling process in its framework according to the experimental results. The experimental results are shown in Table~\ref{tab:readout}. Among all these three pooling functions, min pooling has the worst results on all three datasets. This is because selecting the minimum value may lose some important information. Average pooling shows best performance on DBLP and Yelp, and competitive performance on Aminer, which demonstrates that average pooling has better generalization capability.

\subsubsection{Effect of Embedding Dimension}
\begin{figure}[htbp]
	\centering
	\includegraphics[width=0.5\textwidth]{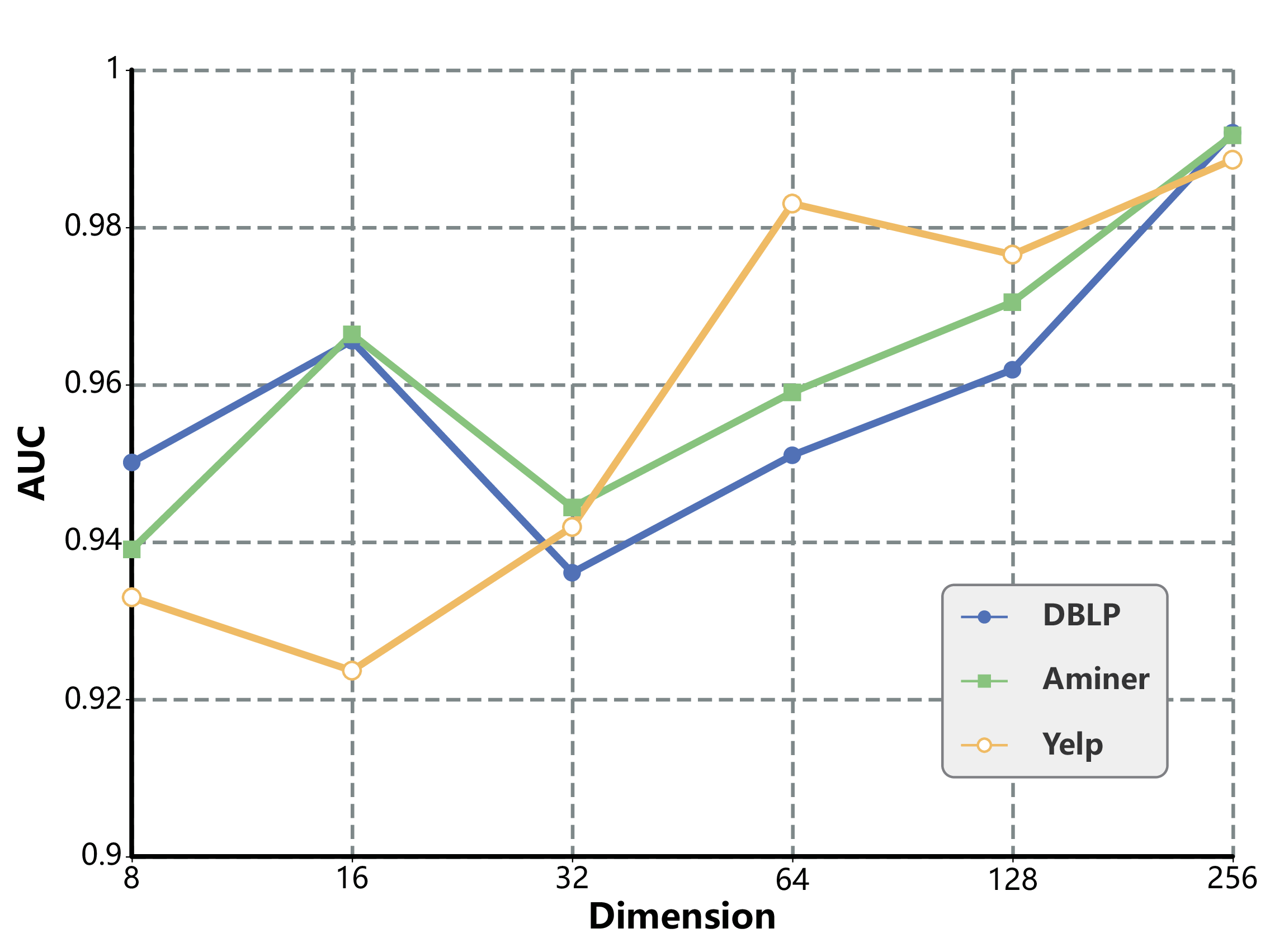}\\
	\caption{The impact of embedding dimension on AUC values.}
	\label{fig:dimension}
\end{figure}

Generally, higher dimensions of embeddings will incorporate more information and have better performance on downstream tasks compared with embeddings with lower dimensions. However, a higher embedding dimension will also use more computation resources and increase the program running time. Therefore, how to balance these two values should be considered carefully by comparing their experiment results, especially in situations having limited resources, which are shown in Figure~\ref{fig:dimension}. In this experiment, we select dimensions ranging from 8 to 256 increased exponentially to compare their corresponding AUC values on three datasets. From this figure, we find that the detection performance does not increase constantly with the increment of embedding dimensions. Despite that 64 dimensions of node embeddings show unsatisfactory results than 128 and 256 dimensions, EAGLE selects to represent node embeddings as 64 for the sake of improving the time efficiency of the model.

\section{Conclusion}~\label{sec6}

In this paper, we propose an efficient anomaly detection model named EAGLE for heterogeneous graphs. By combining the contrastive learning technique and autoencoder module, EAGLE could learn informative node embeddings and identify node anomalies in a self-supervised manner. The instance pairs are sampled from the meta-path level, which captures both the semantic and structural properties. Extensive experiments on three benchmark datasets demonstrate the effectiveness and efficiency of the proposed algorithm. For future work, we plan to investigate dynamic graph learning techniques to detect anomalies in streaming graphs.

%

\bibliographystyle{IEEEtran}
\bibliography{IEEEabrv,mybibfile}

\begin{IEEEbiography}{Jing Ren}{\,}is currently working toward the Ph.D. degree in information technology at Federation University Australia. Her research interests are graph neural networks, anomaly detection, and network science. Contact her at jingr@students.federation.edu.au.
\end{IEEEbiography}

\begin{IEEEbiography}{Mingliang Hou}{\,}received the B.Sc. degree from Dezhou University and the M.Sc. degree from Shandong University, Shandong, China. He is currently pursuing the Ph.D. degree in the School of Software, Dalian University of Technology, Dalian, China. His research interests include graph learning, city science, and social computing.
\end{IEEEbiography}

\begin{IEEEbiography}{Zhixuan Liu}{\,}is currently working toward the M.Sc. degree in software engineering at Dalian University of Technology, Dalian, China. His research interests include automated graph learning and social computing. Contact him at zhixuanliu1998@outlook.com.
\end{IEEEbiography}

\begin{IEEEbiography}{Xiaomei Bai}{\,}received the BSc degree from the University of Science and Technology Liaoning, Anshan, China, in 2000, and the MSc degree from Jilin University, Changchun, China, in 2006. She received Ph.D. degree in School of Software, Dalian University of Technology, Dalian, China in 2017. Since 2000, she has been working in Anshan Normal University, China. Her research interests include computational social science, science of success in science, and big data.
\end{IEEEbiography}

\end{document}